\documentclass[letterpaper]{article} 
\usepackage{aaai2026}  
\usepackage{times}  
\usepackage{helvet}  
\usepackage{courier}  
\usepackage[hyphens]{url}  
\usepackage{graphicx} 
\usepackage{natbib}  
\usepackage{caption} 
\usepackage{amsmath}
\usepackage{adjustbox}   
\usepackage{multirow}    
\usepackage{makecell}    
\usepackage{booktabs}    
\usepackage{amssymb}
\usepackage{makecell}
\usepackage{graphicx}
\usepackage{subcaption}
\usepackage{algorithm}
\usepackage{algorithmic}
\usepackage{adjustbox}
\usepackage{multirow}
\usepackage{makecell}
\usepackage[table,xcdraw]{xcolor}
\usepackage{pifont} 
\usepackage{algorithm}
\usepackage{algorithmic}

\usepackage{newfloat}
\usepackage{listings}
\DeclareCaptionStyle{ruled}{labelfont=normalfont,labelsep=colon,strut=off} 
\floatstyle{ruled}
\newfloat{listing}{tb}{lst}{}
\floatname{listing}{Listing}
\title{Temporal Dynamics Enhancer for Directly Trained Spiking Object Detectors}

\author{
    Fan Luo\textsuperscript{\rm 1,3}, 
    Zeyu Gao\textsuperscript{\rm 1,3}, 
    Xinhao Luo\textsuperscript{\rm 1,3}, 
    Kai Zhao\textsuperscript{\rm 2,3}, 
    Yanfeng Lu\textsuperscript{\rm 1,3}\thanks{Corresponding author.}
}

\affiliations{
    \textsuperscript{\rm 1}State Key Laboratory of Multimodal Artificial Intelligence Systems,\\
    Institute of Automation, Chinese Academy of Sciences (CASIA), Beijing 100190, China\\
    \textsuperscript{\rm 2}Laboratory of Cognition and Decision Intelligence for Complex Systems,\\
    Institute of Automation, Chinese Academy of Sciences (CASIA), Beijing 100190, China\\
    \textsuperscript{\rm 3}School of Artificial Intelligence,\\
    University of Chinese Academy of Sciences, Beijing 100049, China\\
    \{luofan2024, gaozeyu2023, luoxinhao2023, kai.zhao\}@ia.ac.cn,\\
    yanfeng.lv@ia.ac.cn
}

\usepackage{bibentry}

\begin{document}

\maketitle

\begin{abstract}
Spiking Neural Networks (SNNs), with their brain-inspired spatiotemporal dynamics and spike-driven computation, have emerged as promising energy-efficient alternatives to Artificial Neural Networks (ANNs). However, existing SNNs typically replicate inputs directly or aggregate them into frames at fixed intervals. Such strategies lead to neurons receiving nearly identical stimuli across time steps, severely limiting the model's expressive power—particularly in complex tasks like object detection. In this work, we propose the Temporal Dynamics Enhancer (TDE) to strengthen SNNs' capacity for temporal information modeling. TDE consists of two modules: a Spiking Encoder (SE) that generates diverse input stimuli across time steps, and an Attention Gating Module (AGM) that guides the SE generation based on inter-temporal dependencies. Moreover, to eliminate the high-energy multiplication operations introduced by the AGM, we propose a Spike-Driven Attention (SDA) to reduce attention-related energy consumption. Extensive experiments demonstrate that TDE can be seamlessly integrated into existing SNN-based detectors and consistently outperforms state-of-the-art methods, achieving mAP@50-95 scores of 57.7\% on the static PASCAL VOC dataset and 47.6\% on the neuromorphic EvDET200K dataset. In terms of energy consumption, the SDA consumes only 0.240× the energy of conventional attention modules.
\end{abstract}

\begin{links}
    \link{Code}{https://github.com/Mortal825/TDE.git}
\end{links}

\section{Introduction}


As a representative of third-generation neural networks, Spiking Neural Networks (SNNs) \cite{maass1997networks, roy2019towards, zhang2021rectified, guo2023transformer, qu2024spike, zhang2025spike, zhang2025toward1, zhang2025toward2} offer a biologically plausible and energy-efficient alternative, gradually attracting increasing attention. SNNs transmit information through discrete spikes rather than continuous values, significantly reducing data transmission and storage costs. Their inherently asynchronous and event-driven nature further eliminates redundant computation and synchronization overhead. As a result, SNNs achieve high energy efficiency on neuromorphic hardware platforms \cite{poon2011neuromorphic,merolla2014million}.

\begin{figure*}[!t]
    \centering
    \begin{subfigure}{0.48\textwidth}
        \centering
        \includegraphics[width=\textwidth]{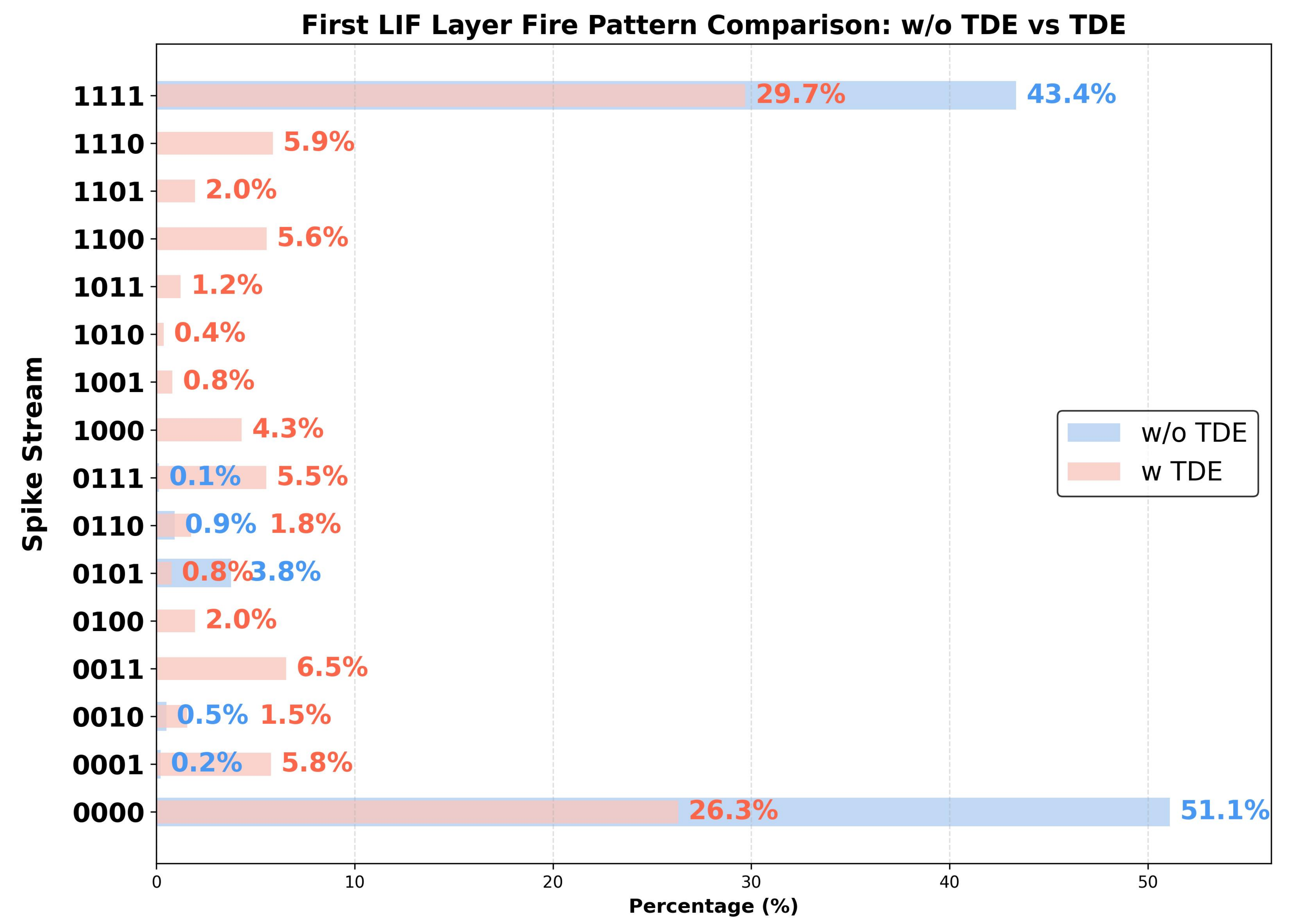}
        \caption{Spike Firing Pattern Before and After the TDE}
        \label{fig:spike_firing_first_layer}
    \end{subfigure}
    \hspace{0.02\textwidth}
    \begin{subfigure}{0.36\textwidth}
        \centering
        \begin{subfigure}{\textwidth}
            \centering
            \includegraphics[width=\textwidth]{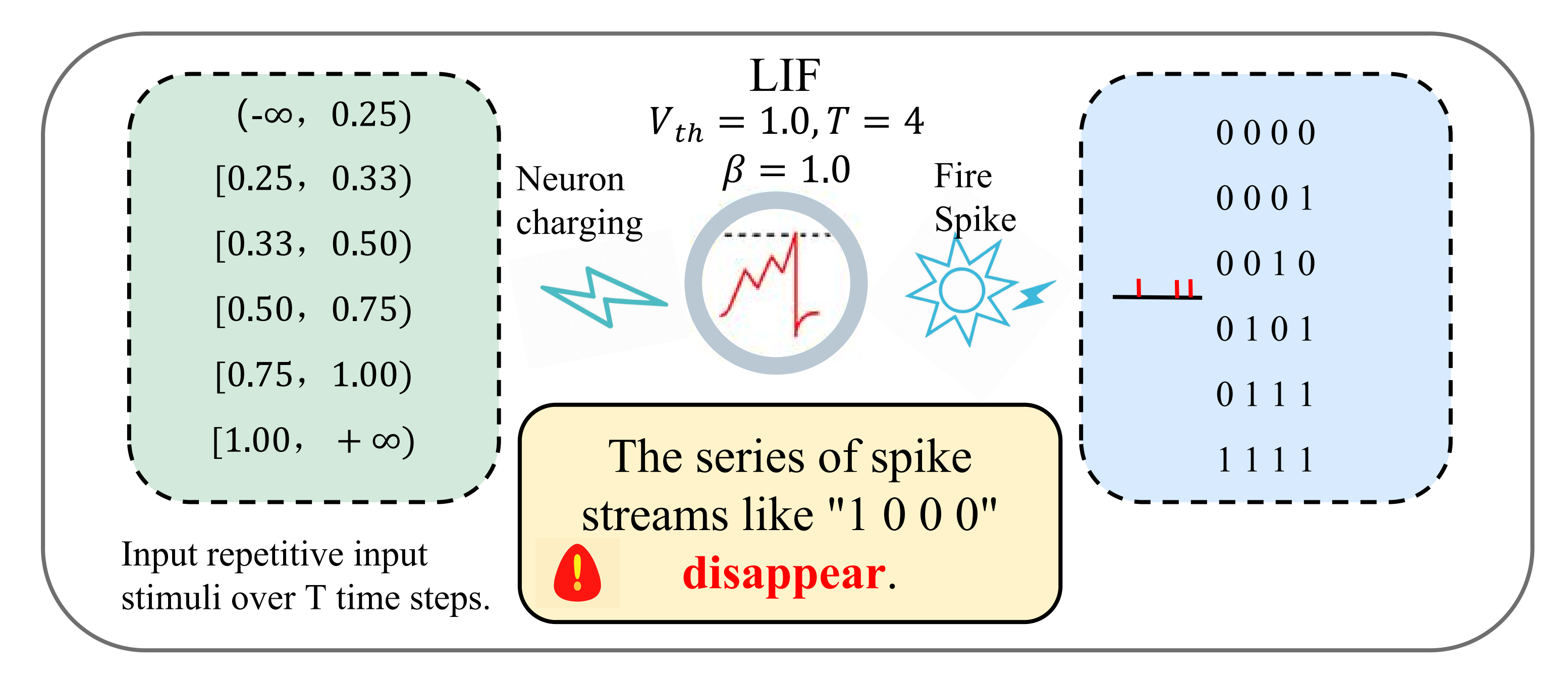}
            \caption{Membrane Potential Accumulation}
            \label{fig:spike_sequence_mapping}
        \end{subfigure}
        \begin{subfigure}{\textwidth}
            \centering
            \includegraphics[width=\textwidth]{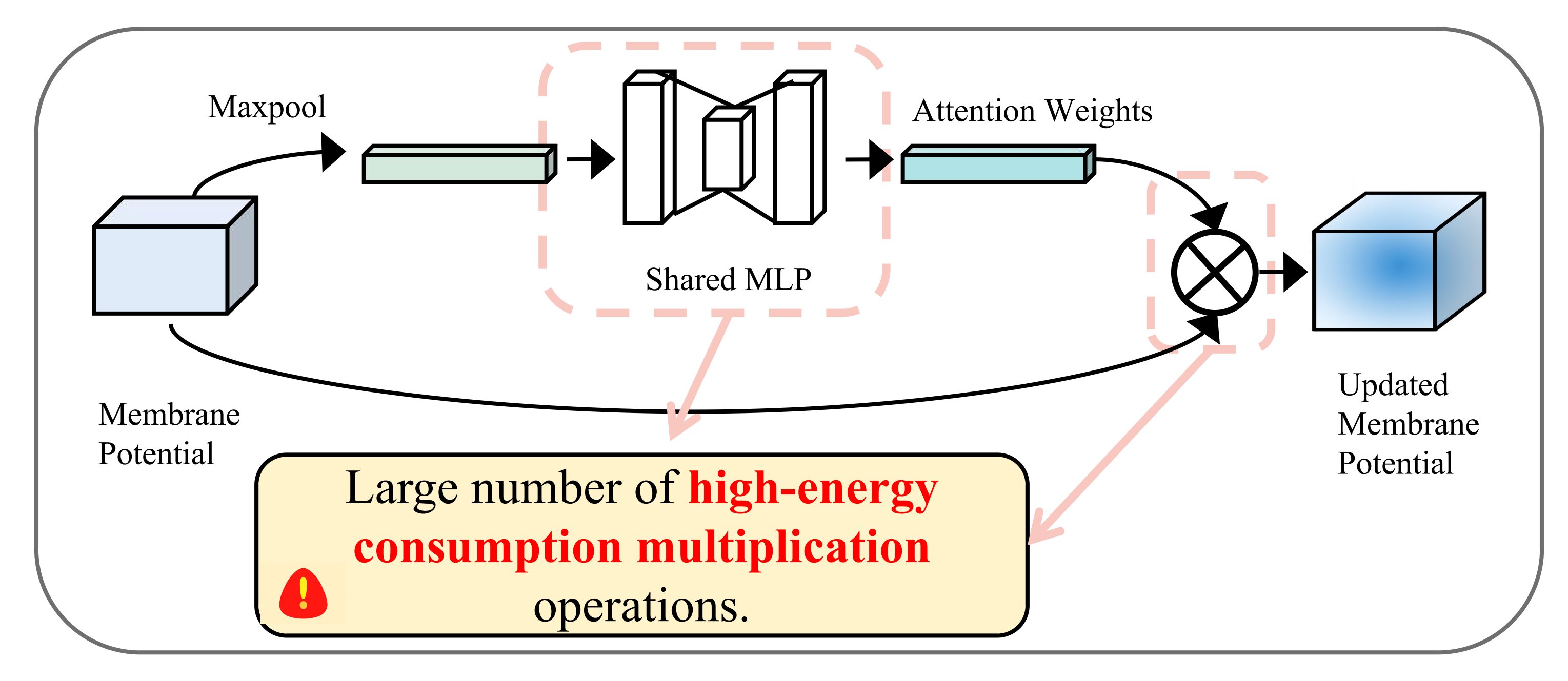}
            \caption{Energy Cost of Traditional Attention Mechanisms in SNNs}
        \end{subfigure}
    \end{subfigure}
    
    \caption{Research Motivation of the TDE Module.
    (a) Illustrates the spike firing pattern of the first LIF neuron layer before and after applying TDE.
    (b) Describes how the LIF neuron in existing SNNs receive repetitive input stimuli, resulting in the disappearance of a series of spike streams.
    (c) Highlights that traditional attention mechanisms in SNNs introduce a large number of energy-intensive multiplication operations.}
    \label{fig:spike_pattern_comparison}
\end{figure*}

Nevertheless, SNNs exhibit limited expressiveness, particularly in complex regression tasks such as object detection. This is primarily due to their spike-based communication paradigm, which inherently restricts the precision and continuity of information representation. Although incorporating multiple time steps enhances temporal expressivity, existing SNN-based detection frameworks still fall short of matching the performance of their ANN counterparts. Recent efforts have attempted to bridge this gap by directly or indirectly incorporating integer spikes \cite{qu2024spiking,luo2024integer,yao2025scaling}. However, direct usage of integer spikes \cite{qu2024spiking} compromises the spike-driven nature of SNNs, while indirect methods \cite{luo2024integer,yao2025scaling} often overlook the hardware costs of converting integer spikes into actual spike events. Importantly, these approaches tend to focus on the representation of single spike information but overlook the temporal domain, which is another critical source of expressive power unique to SNNs. In contrast to ANNs, SNNs possess inherent temporal dynamics, which, if effectively leveraged, could offer a richer and more biologically plausible form of computation.

We argue that existing SNN-based object detection frameworks fail to fully utilize the temporal domain. For static datasets, identical stimuli are repeated across time steps, while for neuromorphic datasets, the long integration window causes similar stimuli, both hindering temporal dynamics. As a result, temporal information is primarily captured through the accumulation of membrane potentials, leaving the broader temporal dimension largely underutilized. As shown in the Fig. \ref{fig:spike_firing_first_layer} and Fig. \ref{fig:spike_sequence_mapping}, the spike streams of the first-layer Leaky Integrate-and-Fire (LIF) neurons exhibit significant disappearance, with spike streams such as ``1110", ``1101". The spike streams tend to be all-silent or all-active, resulting in the temporal dynamics pattern scarcity problem. 


In addition, attention mechanisms have become a key component in SNN frameworks, providing a clear modulation strategy for sparse, temporally-distributed spike representations, allowing the limited expressive power of SNNs to dynamically focus on informative spatial and temporal regions \cite{yao2021temporal,yao2023attention,zhu2024tcja}. However, the non-spiking operations introduced by conventional attention mechanisms contradict the energy-efficient nature of SNNs, severely limiting their practical deployment \cite{qiu2024gated}.

To address the temporal dynamics pattern scarcity problem, we propose \textbf{the Temporal Dynamics Enhancer (TDE)}, a framework that consists of two key components: \textbf{a Spiking Encoder (SE)} that generates temporally diverse input stimuli, and \textbf{an Attention Gating Module (AGM)} that captures inter-temporal dependencies within the network. By introducing temporal diversity and enriching multi-step spike representations, the TDE effectively unlocks the potential of the temporal dimension in SNNs. Moreover, to mitigate the increased high-energy multiplication operations introduced by the attention mechanism in the AGM, we propose \textbf{a Spike-Driven Attention (SDA)} to enable fully accumulation operations reduces attention-related energy consumption.

As shown in Fig. \ref{fig:spike_firing_first_layer}, which illustrates the spike firing pattern of the first LIF layer. After integrating the TDE module, the temporal diversity of fire pattern improves markedly, and all previously missing spike streams in the first-layer LIF neurons are fully recovered.

Our main contributions are summarized as follows:
\begin{itemize}
\item We identify that the reliance on repeated inputs across multiple time steps in SNNs restricts the diversity of spike streams, thereby underutilizing the temporal modeling capabilities of the network.
\item We propose the Temporal Dynamics Enhancer (TDE), a general framework that enhances spike stream diversity and significantly improves the expressiveness of directly trained spiking Object Detectors.
\item We introduce the Spike-Driven Attention (SDA) module, which leverages spatio-temporal spike-based attention to enhance performance while reducing energy consumption to 0.240 times that of non-spiking modules.
\item Through extensive experiments on both static (VOC) and neuromorphic (EvDET200K) datasets, we demonstrate that the Temporal Dynamics Enhancer (TDE) consistently improves performance across a variety of state-of-the-art methods, boosting mAP@50-95 by 1.2\% on the VOC dataset and 2.2\% on the EvDET200K dataset. Our method refreshes the state-of-the-art on both datasets, with the latest mAP@50-95 scores reaching 57.7\% and 47.6\%, respectively.
\end{itemize}

\section{Related Works}
\subsection{Direct training of SNN-based object detectors} 
Direct training of spiking neural network (SNN)-based target detectors has attracted increasing attention, aiming to optimize network architecture and neuronal properties for enhanced performance in object detection tasks. SNNs are typically categorized into CNN-based and Transformer-based models. Owing to their strong inductive bias, CNNs have long dominated the field \cite{d2021convit}, with representative models such as Spiking ResNet \cite{zheng2021going}, SEW-ResNet \cite{fang2021deep}, and MSResNet \cite{hu2024advancing}, differing mainly in LIF neuron placement and identity mapping \cite{he2016identity}. Recently, Transformer-based SNNs \cite{dosovitskiy2020image, vaswani2017attention} have emerged, achieving state-of-the-art results in classification tasks \cite{yao2023spike, zhang2022spike} and enabling the development of directly trained target detectors. Notably, the first such detector was proposed using spike-residue blocks \cite{su2023deep}, and SpikeYOLO \cite{luo2024integer} further enhanced detection performance by incorporating the Meta-SpikeFormer design \cite{yao2024spike}. However, these frameworks still underexplore the inherent temporal dynamics of SNNs and lack explicit temporal modeling.

Many studies \cite{fang2021incorporating,yao2022glif,guo2024ternary} show that binary activation of LIF neurons often results in insufficient precision in complex tasks. Current methods, such as I-LIF neurons \cite{luo2024integer}, address this by using integer values during training and expanding virtual time steps to maintain spike-driven behavior. Similarly, the SFA method \cite{yao2025scaling} uses integer training and spike-driven inference. Ternary spiking neurons \cite{yuan2024trainable,miao2025spikingyolox} improve deep-layer features, enhancing target detection. However, these methods require additional hardware overhead or disrupt the spike-driven nature of SNNs due to their use of non-traditional LIF neurons. We believe the temporal dimension of SNNs holds significant potential, and this paper focuses on exploring this using the binary activation of original LIF neurons.
\subsection{Enhancing Temporal Dynamics in SNNs}
To enhance the temporal dynamics of SNNs, researchers have improved spike representation encoding schemes. For example, Gated Attention Coding (GAC) \cite{qiu2024gated} transforms static images into robust representations with temporal dynamics. Additionally, Frequency Encoding (FE) \cite{xu2024feel} simulates selective visual attention in the biological brain, removing noise at different frequencies. However, these methods have not clearly elucidated the relationship between temporal dynamics and spike firing pattern, and have only been tested on classification tasks with relatively low precision demands.

Many studies \cite{shen2024tim, zhang2025staa, lee2025spiking} have explored the advantages of the temporal dimension when using attention mechanisms in SNNs. Inspired by SE-Net \cite{hu2018squeeze}, researchers extended the attention mechanism to the temporal dimension, evaluating the importance of each frame in the final decision during training \cite{yao2021temporal}, thereby improving the temporal dynamics of SNNs. Modules like TCSA-SNN \cite{yao2023attention} and TCJA-SNN \cite{zhu2024tcja} assess the significance of spike streams across multiple dimensions, further enhancing the temporal dynamics. However, the current temporal-based attention mechanism requires a separate multiplication module to compute attention weights, disrupting the spike-driven nature of SNNs.
\begin{figure*}[!t]
    \centering
    \includegraphics[scale=0.33]{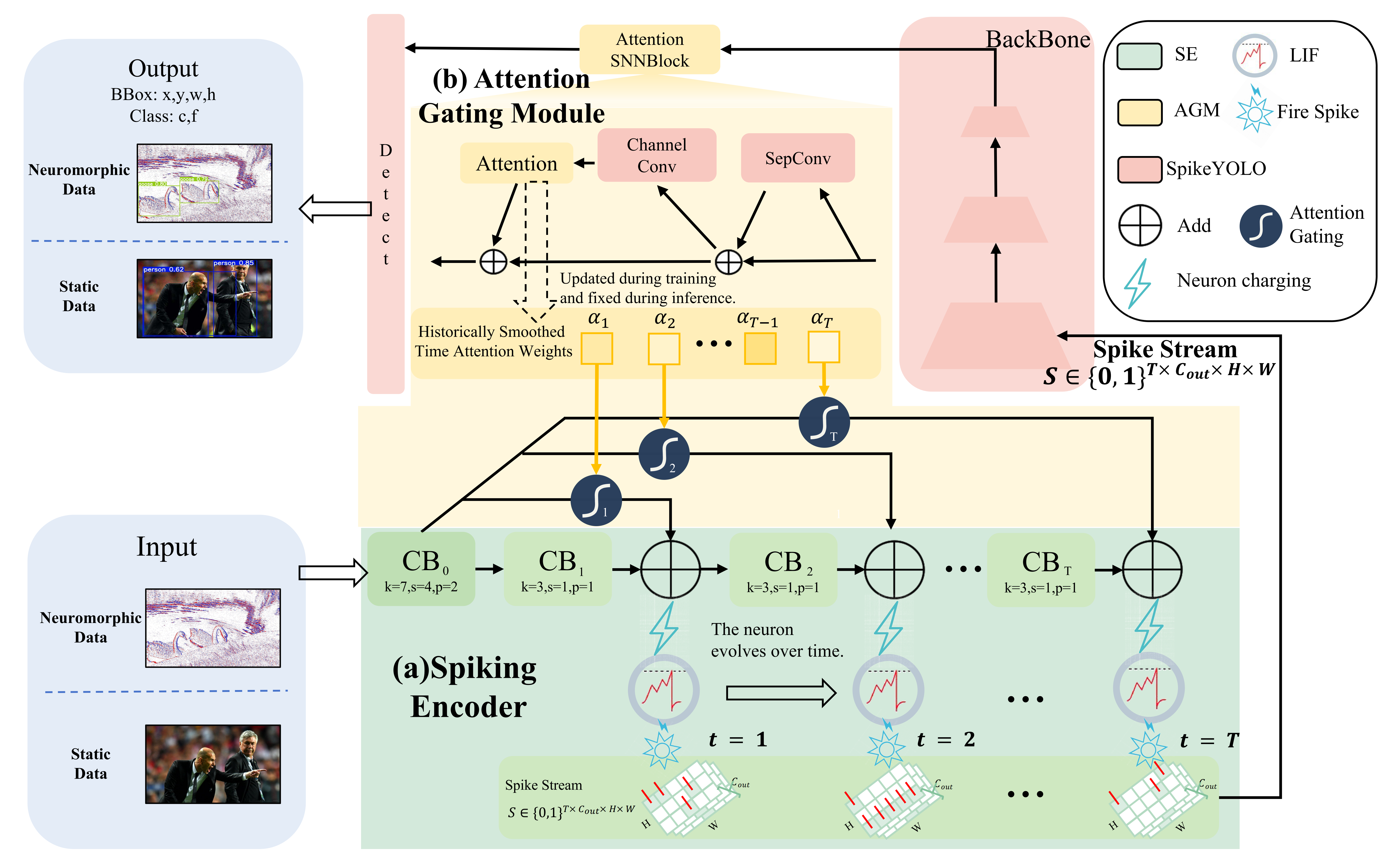}
    \caption{The Temporal Dynamics Enhancer (TDE) consists of two main components: (1) The spiking encoder (SE), using the CB component (Conv-BN), triggers the generation of diverse spikes. Its connection mechanism is based on the charging equation of the LIF neuron and the firing equation. (2) The Attention Gating Module (AGM) enhances the temporal dynamics of neurons within the layer by utilizing a general multi-dimensional attention mechanism. At the same time, it obtains temporal attention weights to regulate the spike stream generation of the SE, suppressing unreasonable exploration.}
    \label{fig:TDE}
\end{figure*}

\section{Preliminary Knowledge}
\textbf{Leaky Integrate-and-Fire neuron (LIF)}: In Spiking Neural Networks, one of the most commonly used neuron models is the Leaky Integrate-and-Fire (LIF) model \cite{tal1997computing}, as it offers a good balance between biological plausibility and computational efficiency. 

Formally, the neuron dynamics at time step \( t \) are given by:
\begin{align}
H_t &= V_{t-1} + X_t, \label{eq:charging} \\
S_t &= \Theta(H_t - V_{th}), \label{eq:firing} \\
V_t &= \beta (H_t - V_{th} S_t), \label{eq:update}
\end{align}
where \( X_t \in \mathbb{R} \) is the input current at time \( t \), \( V_{t-1} \) is the membrane potential carried over from the previous step, and \( H_t \) is the pre-firing potential after integrating input. The neuron fires a spike \( S_t \in \{0, 1\} \) when \( H_t \) exceeds the firing threshold \( V_{th} \), determined by the Heaviside step function \( \Theta(\cdot) \). After firing, the membrane potential is updated using a soft-reset mechanism, where \( \beta \in [0, 1] \) is the leak factor that governs the decay of membrane potential over time.


\textbf{Neuromorphic data}: Neuromorphic data are generated by a Dynamic Vision Sensor (DVS), which triggers events when the logarithmic change in pixel intensity exceeds a threshold. Each event is represented as $(x_n, y_n, t_n, p_n)$, where $(x_n, y_n)$ are spatial coordinates, $t_n$ is the timestamp, and $p_n \in \{-1, 1\}$ indicates polarity (increase or decrease in brightness). A common preprocessing strategy is to accumulate the event streams within a fixed time window into a frame format \cite{yao2023sparser}. The resulting event data can be represented as a tensor of shape $\mathbb{R}^{1 \times H \times W}$, where the single channel encodes the accumulated event polarity information.

\textbf{Direct encoding schemes}: Current SNN-based object detectors often adopt direct encoding \cite{wu2019direct}, where the raw image is injected at each time step and spike streams are implicitly generated by the initial Conv-BN layer.

\section{Method}
In this section, we introduce the Temporal Dynamics Enhancer (TDE) to tackle spike firing pattern scarcity in SNNs for object detection. The TDE comprises two modules: a Spiking Encoder (SE) to generate varied temporal stimuli and an Attention Gating Module (AGM) to mitigate irrational exploration within the SE. By adding additional neuron groups, we achieve a fully Spike-Driven Attention (SDA) without multiplication operations. TDE is a general-purpose module. In this section, we integrate it into the SpikeYOLO framework to illustrate its design and implementation.

\subsection{Spiking Encoder (SE)}
Current SNN-based object detection frameworks predominantly employ direct encoding schemes, where the image input is repeatedly fed into LIF neurons. This repetitive operation fails to generate dynamic output, leading to a lack of temporal dynamics in subsequent SNN architectures. The spiking encoder of an SNN is designed to convert an image input $I \in R^{C_{in} \times H \times W}$ into the spike streams $S \in \{0,1\}^{T \times C_{out} \times H \times W}$. For neuromorphic datasets, we use event-accumulated frames over the entire time window as input.

We first use a convolutional layer to extract initial feature. Considering the membrane potential accumulation property of the LIF neuron model, the feature block at each time step in every $T$ timestep simulations should satisfy:
\begin{equation}
A_{t} =
\begin{cases}
f_{0}(I), & t=0 \\
f_{t}(A_{t-1},I), & 0<t<=T \
\label{eq:time step updata1}
\end{cases}
.
\end{equation}   
Each time step's feature block is influenced not only by external input but also by the feature block of the previous time step. This is consistent with the LIF neuron update process, enabling the generation of spike inputs with rich temporal dynamics. The function $f$ at timestep $t$ is defined as follows:
\begin{equation}
	f_{t}(A_{t-1},I) = \alpha_t I + (1-\alpha_t) f_{t}^{k \times k}(A_{t-1}),
	\label{eq:time step updata2}
\end{equation}
where $\alpha_t$ is the preference coefficient at time step $t$, and $f_{t}^{k \times k}$ denotes the 2D convolution operation at time $t$ with a filter size of $k \times k$.

Subsequently, we obtain a feature sequence $A$. By applying the LIF neuron, we can obtain the spike streams $S$:
\begin{equation}
	S = LIF(A), \quad S \in \{0,1\}^{T \times C_{out} \times H \times W}.
	\label{eq:time step updata3}
\end{equation}

\begin{figure*}[!t]
	\centering
	\includegraphics[scale=0.3]{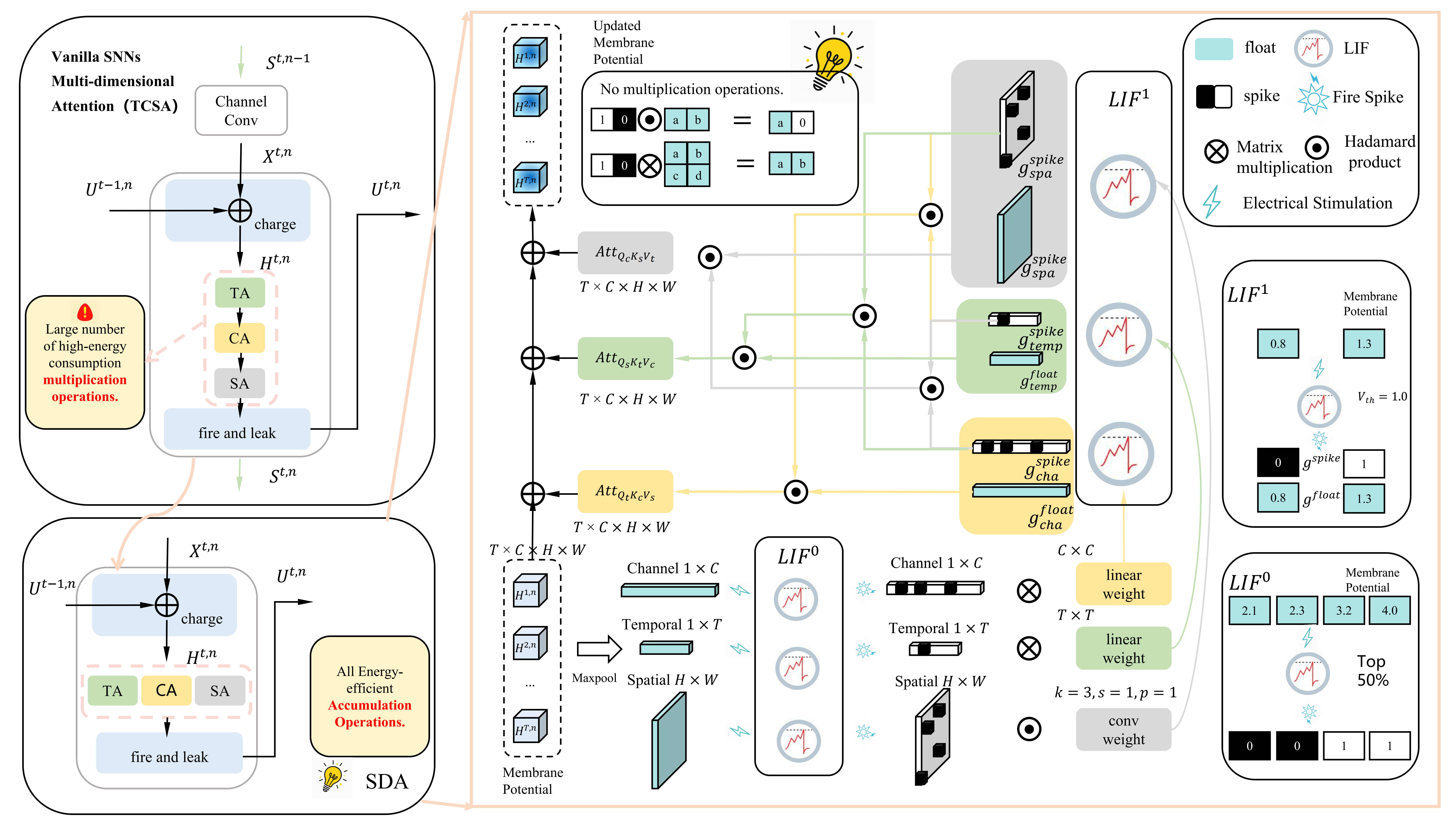}
	\caption{
        SDA: Schematic of the Spike-Driven Attention. SDA uses two neuron groups to avoid the multiplication operations involved. It fuses the spike and floating-point temporal, spatial, and channel attention weights with cross-attention to obtain the attention-updated membrane potential, eliminating the need for hadamard multiplication and matrix multiplication operations in the membrane potential update.
        }
	\label{fig:SDA}
\end{figure*}
\subsection{Attention Gating Module (AGM)}
The Attention Gating Module (AGM) utilizes a general multi-dimensional attention mechanism, which captures temporal attention weights to modulate the output of the SE. This approach, on one hand, leverages the attention mechanism to effectively regulate the membrane potential distribution of the LIF neurons, further enhancing the inter-layer temporal dynamics. On the other hand, by incorporating the attention gating mechanism, we can effectively suppress unreasonable exploration in the SE.

We first integrate a general multi-dimensional attention mechanism as the basic component of the AGM. This general multi-dimensional attention will be introduced in detail later. The computation process for each dimension can be expressed as:
\begin{equation}
H_{Att}^n = g(H^n) \circ H^n,
\label{eq:compute_Hatt}
\end{equation}
where \( g(H^n) \) denotes the function that generates the attention weights, reflecting the process of focusing on discriminative moments or regions. The attention corresponding to each dimension is denoted with a subscript, such as the temporal attention \( g_{\text{temp}}(H^n) \). Here, \( H^n \) typically represents the membrane potential of a neuron in the \( n \)-th layer prior to reset and \( H_{\text{Att}}^n \) denotes the attention-modulated membrane potential.

\begin{algorithm}[t]
\caption{Attention Gating}
\label{alg:attention-gating}
\textbf{Input}: Temporal attention weights $g_{t,\text{temp}}^{\text{float}}$, previous preference coefficient $\bar{\alpha}_{t}$, batch size $B$. \\
\textbf{Output}: Updated preference coefficient ${\alpha}_t$.

\begin{algorithmic}[1]
\FOR{$t = 1$ to $T$}
  \STATE Batch-averaged attention: $\hat{\alpha}_t = \frac{1}{B} \sum_{b=1}^{B} g_{t,b,\text{temp}}^{\text{float}}$
  \STATE Temporal smoothing: ${\alpha}_t = \frac{1}{2}(\bar{\alpha}_{t} + \hat{\alpha}_t),\ \bar{\alpha}_{t} = \alpha_t$
\ENDFOR
\end{algorithmic}
\end{algorithm}

After obtaining temporal attention weights via \( g_{\text{temp}}(H^n) \), the AGM module treats them as structural priors to guide the SE's feature construction and avoid unreasonable exploration. The preference coefficient \(\alpha_t\) in Eq.~\eqref{eq:time step updata2} is updated as shown in Alg.~\ref{alg:attention-gating}. At each time step, the SE fuses current and previous feature blocks based on \(\alpha_t\): higher values emphasize current temporal features, while lower values increase reliance on the original block to suppress unstable or redundant evolution.

\subsection{Spike-Driven Attention (SDA)}
This general multi-dimensional attention has various implementations, such as TCSA \cite{yao2023attention}. However, the current integration of attention blocks requires separate multiplication modules in subsequent layers to dynamically compute attention weights, as seen in the attention weight calculation function \( g(H^n) \) and the membrane potential update with the Hadamard product in Eq. \eqref{eq:compute_Hatt}, which disrupts the spike-driven nature of SNNs. To address this, we propose Spike-Driven Attention (SDA), a multidimensional spiking attention block that enhances temporal dynamics while preserving spike-driven characteristics.

To eliminate multiplication operations, we replace independent multiplication units with a group of neurons. The neuron group consists of two sets, \( LIF^0 \) and \( LIF^1 \). \( LIF^0 \) discards the threshold firing characteristic and uses a top-k\% strategy for spike generation, where neurons receiving the top k\% of stimuli fire, and others remain silent. \( LIF^1 \), in contrast, retains the threshold firing mechanism, but its output also includes the membrane potential, providing an additional representation of the spike signal.

In SNNs, Time Attention (TA) utilizes the temporal relationships of membrane potential to enhance the network's temporal features. To compute the 1-D TA weights, the membrane potential of neurons in the \(n\)-th layer, $H^n = [\dots, H_{\text{temp}}^n, \dots] \in \mathbb{R}^{T \times c_n \times h_n \times w_n}$, are used as input. The 1-D TA spike weights $g_{\text{temp}}^{\text{spike}} \in {\{0,1\}}^{T \times 1 \times 1 \times 1}$ and 1-D TA float weights $g_{\text{temp}}^{\text{float}} \in \mathbb{R}^{T \times 1 \times 1 \times 1}$ can be represented as:

\begin{equation}
g_{\text{temp}}^{\text{spike}}, g_{\text{temp}}^{\text{float}} = LIF_{\text{temp}}^{1}(W_{\text{temp}}^n(LIF_{\text{temp}}^{0}(\text{MaxPool}(H^n)))).
\label{eq:TA}
\end{equation}

Channel Attention (CA) and Spatial Attention (SA) are implemented similarly. Maxpool is used to extract significant regions, LIF$^0$ is used to excite specific regions, and convolution or fully connected layers map these regions. LIF$^1$ is then used to obtain the spike attention weights and float attention weights. A detailed implementation can be found in \textbf{Supplementary Material (Section~I)}.

Finally, we fuse the temporal, spatial, and channel attention weights with cross-attention to obtain the attention-updated membrane potential, \( H_{Att}^n \), in the SDA module as follows:
\begin{equation}
\resizebox{0.95\linewidth}{!}{$
\begin{aligned}
	H_{Att}^n &= g(H^n) + H^n \\
            &= Att_{Q_t-K_c-V_s} + Att_{Q_c-K_s-V_t} + Att_{Q_s-K_t-V_c} + H^n \\
            &= g_{\text{temp}}^{\text{spike}} \times g_{\text{ch}}^{\text{float}} \times g_{\text{spa}}^{\text{spike}} + g_{\text{ch}}^{\text{spike}} \times g_{\text{spa}}^{\text{float}} \times g_{\text{temp}}^{\text{spike}} \\
            &\quad + g_{\text{spa}}^{\text{spike}} \times g_{\text{temp}}^{\text{float}} \times g_{\text{ch}}^{\text{spike}} + H^n.
\end{aligned}
$}
\label{eq:Attention Overview}
\end{equation}

\begin{figure}[!t]
	\centering
	\includegraphics[width=0.42\textwidth]{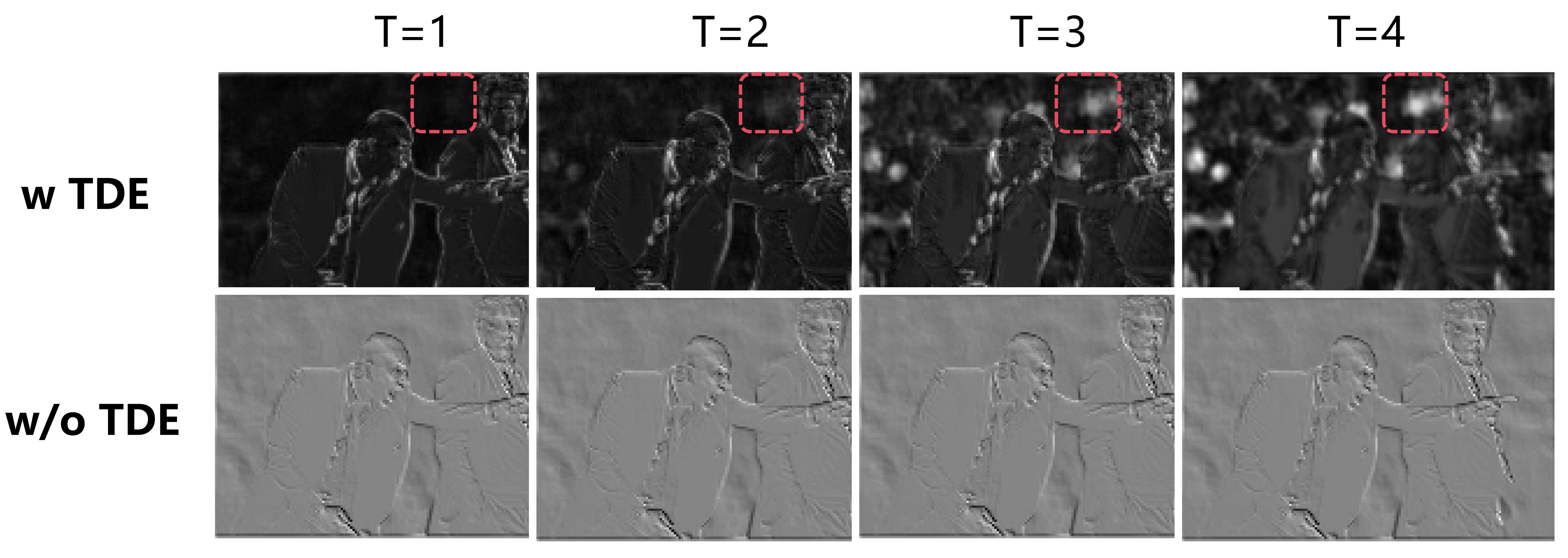}  
	\caption{With TDE, the network gradually shifts attention from the object to surrounding regions over time, while the baseline shows mostly static feature maps.}
	\label{fig:encoder}
\end{figure}

\section{Experiments}
We integrated the TDE module with the three most advanced directly trained SNN-based object detectors (based on binary spikes) to validate its generalization ability. During the experiments, we used consistent hyperparameters and the same experimental device across all methods, without any additional tricks. We set the time step to $T = 4$ to trade-off efficiency and accuracy. More details are provided in \textbf{Supplementary Material (Section~II)}.

\subsection{Dataset and Evaluation Metric}
We validated the generalization ability of the TDE module using state-of-the-art object detectors on the static datasets VOC2007 and VOC2012 \cite{everingham2015pascal}, as well as the neuromorphic dataset EvDET200K \cite{wang2025object}. The VOC2007 and VOC2012 datasets contain 9,963 and 11,530 images, respectively, with annotations for 20 object categories, and are widely used as benchmark datasets for object detection. The EvDET200K dataset consists of 10,054 video streams, covering 10 object categories and providing 202,260 annotations. This dataset focuses on small object detection and includes multiple challenges such as multi-view, multi-illumination, multi-motion, and dynamic backgrounds. For evaluation, we used the mean Average Precision (mAP) at different Intersection over Union (IOU) thresholds, which is the most commonly used metric in object detection. In terms of energy consumption, we calculated the number of floating-point multiplication (MUL) and accumulation (AC) operations for the attention mechanism. Using a 45nm technology node with 32-bit floating-point precision, the energy cost is 3.7 pJ per MUL operation and 0.9 pJ per AC operation \cite{kim2020spiking}. Additionally, we measured the number of parameters for each detector to provide a more comprehensive and accurate understanding of the models' performance.

\subsection{Efficiency and Generalizability Validation}
The TDE (Temporal Dynamics Enhancer) module demonstrates remarkable consistency improvement when integrated with the current state-of-the-art directly trained SNN object detectors (as shown in Tab. \ref{table:3 datasets combined}).

From a global perspective, whether on static datasets like VOC2007 and VOC2012, or on neuromorphic datasets like EvDET200K, integrating the TDE (TCSA) module consistently improves the mAP50 by over 1.1\%. Notably, when combined with SpikeYOLO, the performance on VOC2007 improves by 3.0\%. From a more localized perspective, the integration of TDE enables underperforming methods to surpass others. For example, the original EMSYOLO without TDE achieves a mAP50 of only 44.9\% on EvDET200K, significantly lower than SpikeYOLO’s 46.5\%. However, with the addition of the TDE module, EMSYOLO’s performance on EvDET200K improves to 47.1\%, surpassing other frameworks. A similar trend is observed on VOC2007, where the original EMSYOLO without TDE has a mAP50 of 31.9\%, well below SpikeYOLO’s 33.2\%. After integrating TDE, EMSYOLO’s performance on VOC2007 increases to 34.9\%, achieving a similar leap. When TDE is integrated into Meta-SpikeFormer, consistent performance improvements are observed. However, its performance on the VOC dataset remains low, likely due to the lack of CNN's prior inductive bias in Meta-SpikeFormer. It is also worth mentioning that integrating the TDE module into the three directly trained SNN object detectors increases the parameter count by no more than 0.26M, further highlighting the efficiency of the TDE module. Fig. \ref{fig:encoder} qualitatively illustrates how TDE enhances the temporal dynamics of SNNs.

To reduce energy consumption, we replaced the non-spiking attention TCSA module with the SDA module, eliminating all multiplication operations. As a result, SDA's energy consumption is only 0.240 times that of TCSA (see Tab. \ref{table:energy_voc2007}). While reducing energy consumption, TDE(SDA) still demonstrates continuous performance improvements. Specifically, when tested with the SpikeYOLO method on the VOC2007 dataset, mAP@50 reached 35.6\%, surpassing the 34.9\% achieved by the non-spiking version TDE(TCSA). However, on the EvDET200K dataset, the performance improvement with TDE(SDA) was lower than that of TDE(TCSA). We believe this is due to the limited learning capacity of the SDA Spike-Driven Attention, as it struggles to capture the intrinsic relationships in data flows (such as time, space, and channels) when working with larger datasets. Nonetheless, its sparse attention mechanism helps prevent overfitting on smaller datasets.

\begin{table}[t]
\caption{Energy analysis of TDE (TCSA vs. SDA) attention mechanisms in SpikeYOLO on VOC2007.}
\label{table:energy_voc2007}
\centering
\begin{adjustbox}{max width=0.9\linewidth}
\begin{tabular}{lcccc}
\toprule
\textbf{Method} & \textbf{MUL} & \textbf{AC} & \textbf{Energy ($\mu J$)} & \textbf{Energy Ratio} \\
\midrule
TDE(TCSA)  & $5.75 \, \text{E}6$ & $5.76 \, \text{E}5$ & $2.18 \, \text{E}1$    & $1.0$ \\
TDE(SDA)   & 0 & $5.82 \, \text{E}6$ & $5.24 \, \text{E}0$   & $\textbf{0.240}$ \\
\bottomrule
\end{tabular}
\end{adjustbox}

\begin{minipage}{0.9\linewidth}
\footnotesize \textit{Note:} Energy computed for attention over membrane potentials with $T{=}4$, $C{=}128$, $H{=}80$, $W{=}40$.
\end{minipage}
\end{table}


\begin{table*}[t]
\caption{Performance improvement of spiking object detection models with TDE on different datasets.}
\label{table:3 datasets combined}
\centering
\begin{adjustbox}{max width=\linewidth}
\begin{tabular}{ccccccc}
\toprule
Dataset & Methods & Models & \makecell{Params\\(M)} & \makecell{Time\\Steps} & mAP@50 (\%) & mAP@50:95 (\%) \\
\midrule
\multirow{9}{*}{VOC} 
& ANN2SNN  & SUHD\cite{qu2024spiking} & - & 4 & 75.3 & - \\
& \multirow{8}{*}{Direct training} 
& SpikeYOLO\cite{luo2024integer} & 23.156 & 4 & 78.0 & 56.6 \\
&& \cellcolor{gray!40}+TDE (TCSA) & \cellcolor{gray!40}23.416 & \cellcolor{gray!40}4 & \cellcolor{gray!40}79.1 (+1.1) & \cellcolor{gray!40}57.7 (+1.1) \\
&& \cellcolor{gray!20}+TDE (SDA) & \cellcolor{gray!20}23.645 & \cellcolor{gray!20}4 & \cellcolor{gray!20}78.3 (+0.3) & \cellcolor{gray!20}57.1 (+0.5) \\
\cmidrule(lr){3-7}
&& EMSYOLO\cite{su2023deep} & 33.889 & 4 & 76.8 & 49.6 \\
&& \cellcolor{gray!40}+TDE (TCSA) & \cellcolor{gray!40}34.126 & \cellcolor{gray!40}4 & \cellcolor{gray!40}77.1 (+0.3) & \cellcolor{gray!40}50.8 (+1.2) \\
&& \cellcolor{gray!20}+TDE (SDA) & \cellcolor{gray!20}34.358 & \cellcolor{gray!20}4 & \cellcolor{gray!20}77.3 (+0.5) & \cellcolor{gray!20}50.2 (+0.6) \\
\cmidrule(lr){3-7}
&& Meta-SpikeFormer\cite{yao2024spike} & 16.652 & 4 & 49.8 & 24.3 \\
&& \cellcolor{gray!40}+TDE (TCSA) & \cellcolor{gray!40}16.823 & \cellcolor{gray!40}4 & \cellcolor{gray!40}51.7 (+1.9) & \cellcolor{gray!40}25.7 (+1.4) \\
\midrule
\multirow{9}{*}{EvDET200K} 
& \multirow{9}{*}{Direct training} 
&  SpikeYOLO\cite{luo2024integer} &  23.156  &  4 &  75.2 (\textit{74.8}) &  46.5 (\textit{41.2}) \\
&& \cellcolor{gray!40}+TDE (TCSA) & \cellcolor{gray!40}23.416 & \cellcolor{gray!40}4 & \cellcolor{gray!40}76.0 (+0.8) & \cellcolor{gray!40}47.6 (+1.1) \\
&& \cellcolor{gray!20}+TDE (SDA) & \cellcolor{gray!20}23.645 & \cellcolor{gray!20}4 & \cellcolor{gray!20}75.8 (+0.6) & \cellcolor{gray!20}47.2 (+0.7) \\
\cmidrule(lr){3-7}
&& EMSYOLO\cite{su2023deep} &  33.889 & 4 &  77.2 (\textit{66.6}) & 44.9 (\textit{32.1}) \\
&& \cellcolor{gray!40}+TDE (TCSA) & \cellcolor{gray!40}34.126  & \cellcolor{gray!40}4 & \cellcolor{gray!40}78.2 (+1.0) & \cellcolor{gray!40}47.1 (+2.2) \\
&& \cellcolor{gray!20}+TDE (SDA) & \cellcolor{gray!20}34.358  & \cellcolor{gray!20}4 & \cellcolor{gray!20}77.7 (+0.5) & \cellcolor{gray!20}45.9 (+1.0) \\
\cmidrule(lr){3-7}
&& Meta-SpikeFormer\cite{yao2024spike} & 16.652 & 4 & 75.2 &  45.9 \\
&& \cellcolor{gray!40}+TDE (TCSA) & \cellcolor{gray!40}16.823 & \cellcolor{gray!40}4 & \cellcolor{gray!40}76.3 (+1.1) & \cellcolor{gray!40}47.2 (+1.3) \\
\midrule
\multirow{7}{*}{VOC2007} 
& \multirow{7}{*}{Direct training} 
& SpikeYOLO\cite{luo2024integer} & 23.156 &  4 &  51.7 &  31.9 \\
&& \cellcolor{gray!40}+TDE (TCSA) & \cellcolor{gray!40}23.416 & \cellcolor{gray!40}4 & \cellcolor{gray!40}55.9 (+4.2) & \cellcolor{gray!40}34.9 (+3.0) \\
&& \cellcolor{gray!20}+TDE (SDA) & \cellcolor{gray!20}23.645 & \cellcolor{gray!20}4 & \cellcolor{gray!20}56.2 (+4.5) & \cellcolor{gray!20}35.6 (+3.7) \\
\cmidrule(lr){3-7}
&& EMSYOLO\cite{su2023deep} &  33.889 &  4 &  59.8 &  33.2 \\
&& \cellcolor{gray!40}+TDE (TCSA) & \cellcolor{gray!40}34.126 & \cellcolor{gray!40}4 & \cellcolor{gray!40}61.3 (+1.5) & \cellcolor{gray!40}35.1 (+1.9) \\
&& \cellcolor{gray!20}+TDE (SDA) & \cellcolor{gray!20}\cellcolor{gray!20}34.358 & \cellcolor{gray!20}4 & \cellcolor{gray!20}61.3 (+1.5) & \cellcolor{gray!20}34.3 (+1.1) \\
\bottomrule
\end{tabular}
\end{adjustbox}
\noindent\footnotesize{\textit{Note: Numbers in parentheses (in \textit{italic}) indicate baseline results directly quoted from \cite{wang2025object} as reference values.The VOC in the table refers to the VOC2007 and VOC2012 datasets.}}
\end{table*}

\begin{table}[t]
\caption{Ablation study of the proposed modules on the VOC2007 dataset.}
\label{table:ablation}
\centering
\begin{adjustbox}{max width=0.9\linewidth}
\begin{tabular}{cccccc}
\toprule
\textbf{Architecture} & \textbf{AGM} & \textbf{SE} & \textbf{mAP@50 (\%)} & \textbf{mAP@50:95 (\%)} \\
\midrule
\multirow{6}{*}{SpikeYOLO} 
& \ding{55} & \ding{55} & 51.7 & 31.9 \\
& TCSA & \ding{55} & 54.3 {\textbf{(+2.6)}} & 33.9 {\textbf{(+2.0)}} \\
& SDA & \ding{55} & 52.6 {\textbf{(+0.9)}} & 32.6 {\textbf{(+0.7)}} \\
& \ding{55} & \ding{51} & 55.2 {\textbf{(+3.5)}} & 34.6 {\textbf{(+2.7)}} \\
& TCSA & \ding{51} & 55.9 {\textbf{(+4.2)}} & 35.1 {\textbf{(+3.2)}} \\
& SDA & \ding{51} & 56.2 {\textbf{(+4.5)}} & 35.6 {\textbf{(+3.7)}} \\
\midrule
\multirow{6}{*}{EMSYOLO} 
& \ding{55} & \ding{55} & 59.8 & 33.2 \\
& TCSA & \ding{55} & 60.2 {\textbf{(+0.4)}} & 33.9 {\textbf{(+0.7)}} \\
& SDA & \ding{55} & 60.6 {\textbf{(+0.8)}} & 33.8 {\textbf{(+0.6)}} \\
& \ding{55} & \ding{51} & 60.9 {\textbf{(+1.1)}} & 34.1 {\textbf{(+0.9)}} \\
& TCSA & \ding{51} & 61.3 {\textbf{(+1.5)}} & 35.1 {\textbf{(+1.9)}} \\
& SDA & \ding{51} & 61.3 {\textbf{(+1.5)}} & 34.3 {\textbf{(+1.1)}} \\
\bottomrule
\end{tabular}
\end{adjustbox}

\footnotesize{\textit{Note: When AGM is tested alone, the gating mechanism is absent, reducing it to a simple multi-dimensional attention.}}
\end{table}

\subsection{Component Analysis and Ablation Study}
To validate the effectiveness of the proposed components, we conducted an ablation study on the SpikeYOLO and EMSYOLO methods using the VOC2007 dataset (see Tab. \ref{table:ablation}). It is important to note that when AGM is tested in isolation, the gating mechanism is absent, reducing it to a simple multi-dimensional attention mechanism.

Both TCSA and SDA consistently outperform the baseline, supporting prior research. The attention mechanism improves performance by focusing on salient regions and ignoring redundant information, while the SE module generates high-temporal dynamic pulses, addressing the issue of disappearing pulse patterns in traditional object detection, enhancing the expressiveness of binary SNNs. In terms of mAP@50-95, SpikeYOLO improved by 2.7\%, and EMSYOLO by 0.9\%, outperforming TCSA by 2.0\% and 0.7\%, respectively. 
By using Attention Gating to integrate SE and multi-dimensional attention, performance improves consistently, with SpikeYOLO showing a 3.7\% increase and EMSYOLO a 1.9\% increase. This highlights the effective coupling of the two components through the attention gating mechanism. To further assess its impact, an  experiment was conducted on the SpikeYOLO framework (see Tab. \ref{tab:att_gate}), resulting in a 0.7\% boost in mAP@50-95. More results can be found in \textbf{Supplementary Material (Section~III)}.

\begin{table}[t]
\caption{Effect of the Attention Gating (Alg.~\ref{alg:attention-gating}) on the SpikeYOLO Framework for VOC Datasets}
\label{table:agm_attention_spikeyolo_voc}
\centering
\begin{adjustbox}{max width=0.9\linewidth}
\label{tab:att_gate}
\begin{tabular}{lcc}
\toprule
\textbf{Method} & \textbf{mAP@50 (\%)} & \textbf{mAP@50:95 (\%)} \\
\midrule
TCSA + SE        & 78.5 & 57.0 \\
AGM(TCSA) + SE   & 79.1 (+0.6) & 57.7 (+0.7) \\
\bottomrule
\end{tabular}
\end{adjustbox}
\end{table}

\section{Conclusion}
We propose the Temporal Dynamics Enhancer (TDE) to address the limitations of Spiking Neural Networks (SNNs) in temporal modeling for object detection. Through extensive experiments on both static (VOC) and neuromorphic (EvDET200K) datasets, we demonstrate that TDE consistently enhances performance across a variety of state-of-the-art methods. We hope our research provides new insights into the potential of temporal dynamics in SNNs and inspires future research on more efficient and biologically inspired spike-driven learning paradigms.
\clearpage
\section{Acknowledgment}
This work is supported by the Strategic Priority Research Program of the Chinese Academy of Sciences under (Grants XDA0450000, XDA0450202), and Central Government Guidance Funds for Local Science and Technology Development (YDZX2025124) \bibliography{arxiv}
\input{ReproducibilityChecklist.tex}

\end{document}